\documentclass[nolineno]{ieeeaccess}
\usepackage{cite}
\usepackage{amsmath,amssymb,amsfonts}
\usepackage{algorithmic}
\usepackage{graphicx}
\usepackage{textcomp}

\usepackage{bm}

\usepackage{booktabs}
\usepackage{multirow} 
\usepackage{adjustbox}
\usepackage{algorithm}
\usepackage{algorithmic}
\usepackage[normalem]{ulem}
\usepackage{makecell}
\usepackage{caption}
\usepackage{placeins}
 \usepackage{hyperref} 
\usepackage{url}

\usepackage{soul}
\sethlcolor{yellow}

\newif\ifshowchanges
\showchangesfalse    

\newcommand{\rev}[1]{%
  \ifshowchanges
    \hl{#1}%
  \else
    #1%
  \fi
}

\makeatletter
\AtBeginDocument{\DeclareMathVersion{bold}
\SetSymbolFont{operators}{bold}{T1}{times}{b}{n}
\SetSymbolFont{NewLetters}{bold}{T1}{times}{b}{it}
\SetMathAlphabet{\mathrm}{bold}{T1}{times}{b}{n}
\SetMathAlphabet{\mathit}{bold}{T1}{times}{b}{it}
\SetMathAlphabet{\mathbf}{bold}{T1}{times}{b}{n}
\SetMathAlphabet{\mathtt}{bold}{OT1}{pcr}{b}{n}
\SetSymbolFont{symbols}{bold}{OMS}{cmsy}{b}{n}
\renewcommand\boldmath{\@nomath\boldmath\mathversion{bold}}}
\makeatother

\def\BibTeX{{\rm B\kern-.05em{\sc i\kern-.025em b}\kern-.08em
    T\kern-.1667em\lower.7ex\hbox{E}\kern-.125emX}}

\usepackage{etoolbox}

\makeatletter
\patchcmd{\@maketitle}
  {\doifont Digital Object Identifier\space\@doi}
  {}
  {}
  {\PackageWarning{arxiv}{DOI patch failed; check ieeeaccess.cls}}
\makeatother

\begin{document}
\history{}
\doi{}

\title{SAP-ViT: A Self-Attributing Patch Vision Transformer for Medical Image Classification}
\author{\uppercase{Wei Zhang},
\uppercase{Wolfgang Nejdl},
\uppercase{Ming Tang}
}

\address{L3S Research Center, Leibniz University Hannover, 30167 Hannover, Germany}

\markboth
{W. Zhang \headeretal: SAP-ViT: A Self-Attributing Patch Vision Transformer for Medical Image Classification}
{W. Zhang \headeretal: SAP-ViT: A Self-Attributing Patch Vision Transformer for Medical Image Classification}

\corresp{Corresponding author: Wei Zhang (wei.zhang@l3s.de).}

\begin{abstract}
Vision Transformer (ViT) classifiers commonly rely on a global \rev{classification (CLS) token} for image-level prediction, which is poorly aligned with local interpretability requirements in medical imaging. We propose SAP-ViT, a Self-Attributing Patch Vision Transformer that replaces CLS-based classification with direct aggregation of last-layer patch embeddings. This design makes class predictions decomposable into patch-level contributions as part of the forward pass. To encourage concise and readable evidence, SAP-ViT applies sparsity regularization to the patch-weight distribution. Since self-attention mixes information across patch tokens, we further introduce an input-space back-mapping mechanism to estimate which input patches contribute to influential output patch embeddings.
We evaluate SAP-ViT qualitatively and quantitatively on multiple medical image classification datasets spanning different imaging modalities. The results show that SAP-ViT maintains strong classification performance while providing consistent and interpretable patch-level explanations across datasets. 
Ablation results further show that both sparsity regularization and the alignment component of input-space back-mapping contribute to the final explanation quality.
Overall, SAP-ViT provides a favorable trade-off between predictive accuracy and interpretability for ViT-based medical image classification.

\end{abstract}


\titlepgskip=-21pt

\maketitle
\pagestyle{plain}
\thispagestyle{plain}

\section{Introduction}

\rev{Transformer-based models have recently been explored across diverse medical imaging tasks}\cite{ref_khawaji2025}. In medical image classification, models are increasingly expected not only to make accurate predictions but also to provide interpretable evidence for those predictions. This requirement is especially important in clinical settings, where model outputs may support high-stakes decisions and where clinicians need to inspect whether the model relies on medically meaningful image regions.


Vision Transformer (ViT) \cite{ref_vit} classifiers have shown strong performance in image classification, but standard ViT classifiers usually rely on a global \rev{classification (CLS) token} for prediction. The CLS token aggregates information across patch tokens and produces an image-level representation, which is effective for classification but poorly aligned with local interpretability. In medical imaging, where disease evidence is often spatially localized, it is desirable to understand how individual image regions contribute to the prediction.

A natural solution is to make the prediction itself patch-decomposable. However, patch-level attribution in ViT is non-trivial because self-attention \cite{ref_transformer} mixes information across patch tokens. As a result, an influential last-layer patch embedding may contain information not only from its corresponding input patch but also from other regions of the image. Therefore, patch-level attribution in ViT requires both a decomposable prediction mechanism and a way to relate influential output patch embeddings back to input-space evidence.

In this work, we propose SAP-ViT, a Self-Attributing Patch Vision Transformer for medical image classification. SAP-ViT replaces CLS-based classification with direct aggregation of last-layer patch embeddings. 
Each patch embedding is assigned a learned aggregation weight, allowing the class prediction to be decomposed into patch-level contributions as part of the forward pass. 
We use the term self-attributing because attribution is built into the prediction mechanism itself, rather than being produced only by an external post-hoc explainer.

To encourage concise and readable explanations, SAP-ViT applies sparsity regularization to the patch-weight distribution. This encourages the model to rely on a compact set of informative patches rather than distributing evidence diffusely across the entire image. Since self-attention mixes information across tokens, SAP-ViT further introduces an input-space back-mapping mechanism. This module estimates which input patches contribute to influential output patch embeddings, providing an approximate input-space attribution of the model’s patch-level decision evidence.

Our contributions are threefold:

First, we propose a patch-additive ViT classifier that predicts directly
from last-layer patch embeddings rather than the final CLS representation,
enabling explicit decomposition of class predictions into patch-level
contributions.

Second, we introduce sparsity regularization on patch weights to encourage the model to rely on a compact set of informative image regions.

Third, we propose an input-space back-mapping mechanism that estimates which input patches contribute to influential output patch embeddings, addressing the token-mixing effect of self-attention.

\section{Related Work}

\subsection{Post-hoc explanation methods}

\rev{Post-hoc explanation methods explain a trained classifier without modifying its prediction mechanism. Gradient-based methods estimate importance from the sensitivity of a class score to the input or an intermediate representation. Vanilla gradient saliency directly differentiates the class score with respect to the input, integrated Gradients accumulates gradients along a path from a reference input, SmoothGrad averages gradients over perturbed inputs, and Gradient-weighted Class Activation Mapping (Grad-CAM) combines gradients with intermediate activations to produce localization maps}\cite{ref_saliency,ref_ig,ref_smoothgrad,ref_grad_cam}\rev{. More recently, LeGrad adapts gradient-based explanation to ViTs by estimating patch relevance through feature-formation sensitivity}\cite{ref_legrad}\rev{. Although these approaches do not require changes to the model architecture, their explanations may be noisy, sensitive to the selected baseline or layer, or affected by gradient saturation.}

\rev{Perturbation-based methods instead estimate importance by measuring how predictions change when parts of the input are modified. Representative approaches include occlusion analysis, Randomized Input Sampling for Explanation (RISE), meaningful perturbation, differentiable patch masking for ViTs, Local Interpretable Model-agnostic Explanations (LIME), and SHapley Additive exPlanations (SHAP)}\cite{ref_occlusion,ref_rise,ref_pert_1,ref_pert_2,ref_lime,ref_shap}\rev{. Perturbation provides a direct way to examine the model response to selected image regions, but it can be computationally expensive and sensitive to the masking, sampling, superpixel, or reference-distribution strategy. Moreover, heavily perturbed images may depart from the distribution on which the classifier was trained.}

\rev{For ViTs, attention-based methods exploit the information available in the Transformer architecture. Attention rollout estimates information flow by aggregating attention matrices across layers}\cite{ref_att_rollout}\rev{, while transformer attribution and attention-aware relevance propagation combine attention with gradients or relevance-propagation rules to produce class-specific explanations}\cite{ref_re_1,ref_re_2}\rev{. However, attention can represent information routing rather than contribution to the prediction}\cite{ref_attn_not_expl_1,ref_attn_not_expl_2}\rev{. More generally, these methods produce explanations after the prediction has been computed, whereas SAP-ViT incorporates a patch-level decomposition into its prediction mechanism.}

\subsection{Interpretable-by-design models}

\rev{Interpretable-by-design models construct predictions from components whose contributions can be inspected directly. BagNet predicts by aggregating evidence from local image regions, making it possible to examine the spatial evidence supporting the output}\cite{ref_bagnet}\rev{. Neural Additive Models extend additive prediction by learning a nonlinear contribution function for each input feature, while Sum-of-Parts constructs predictions from learned feature groups}\cite{ref_nam,ref_sop}\rev{. These approaches provide important foundations for decomposable prediction, but they are not specifically designed for contextualized ViT patch embeddings, where self-attention repeatedly mixes information across image regions.}

\rev{Several architectural approaches have been developed specifically for ViT interpretation. The Interpretability-Aware Vision Transformer (IA-ViT) jointly trains an interpreter and predictor to generate an explanation alongside the prediction}\cite{ref_ia_1}\rev{. Disentangling Visual Transformers aims to make patch representations more interpretable by reducing token entanglement}\cite{ref_ia_2}\rev{, while Prompt-CAM introduces class-specific prompts and uses prompt attention to localize discriminative regions}\cite{ref_ia_3}\rev{. These methods improve ViT interpretability through a separate interpreter, modified token representations, or prompt attention. In contrast, SAP-ViT uses a linear classification head so that each class logit can be decomposed into patch-specific terms. The contribution to a particular class is jointly determined by the learned patch weight, the corresponding patch embedding, and the class-specific parameters of the classification head.}

\rev{Recent work has also investigated patch selection for regularizing and interpreting ViTs in medical imaging. Nilsson and Azizpour learn stochastic token selection for echocardiography video analysis and left ventricular ejection fraction prediction, demonstrating the relevance of patch-level selection in medical applications}\cite{ref_medical_patch_selection}\rev{. Their approach interprets the model through the selected patch tokens, whereas SAP-ViT focuses on decomposing class predictions through contextualized last-layer patch embeddings and approximately mapping this evidence back to the input space.}

\subsection{Interpretability in V\lowercase{i}T}

\rev{Two properties of ViTs are particularly important for local interpretation. First, a CLS token is commonly used as the global representation for classification. Through self-attention, it interacts with the patch tokens and aggregates information from the image, making a CLS-based prediction difficult to decompose into local contributions}\cite{ref_vit,ref_cls_patch}\rev{. Second, self-attention repeatedly mixes information across patch tokens. Consequently, a last-layer patch embedding may contain information from several input regions rather than only its spatially corresponding input patch. The emergence of high-norm or special-purpose tokens further demonstrates that token representations should not automatically be interpreted as direct spatial evidence}\cite{ref_register}\rev{. Therefore, the location of an influential output patch token does not necessarily identify the input region responsible for its contribution.}

\rev{Existing attention- and relevance-based methods attempt to trace information flow through the Transformer}\cite{ref_att_rollout,ref_re_1,ref_re_2}\rev{, but they do not make the class prediction itself additive over patch-level evidence. Conversely, weighting the last-layer patch embeddings and applying a linear classification head produces a decomposable class prediction, but the resulting output embeddings remain contextualized through self-attention. Their weights therefore do not by themselves establish which input regions supplied the information used by each output embedding. This leaves a specific interpretability gap: connecting patch-additive prediction to input-space evidence while explicitly accounting for token mixing. SAP-ViT addresses this gap through patch-additive prediction, sparsity regularization, and input-space back-mapping. The back-mapping provides an approximate association between contextualized output evidence and input patches rather than guaranteed recovery of causal input contributions.}


\begin{figure*}[!t]
    \centering
    \includegraphics[width=0.9\textwidth]{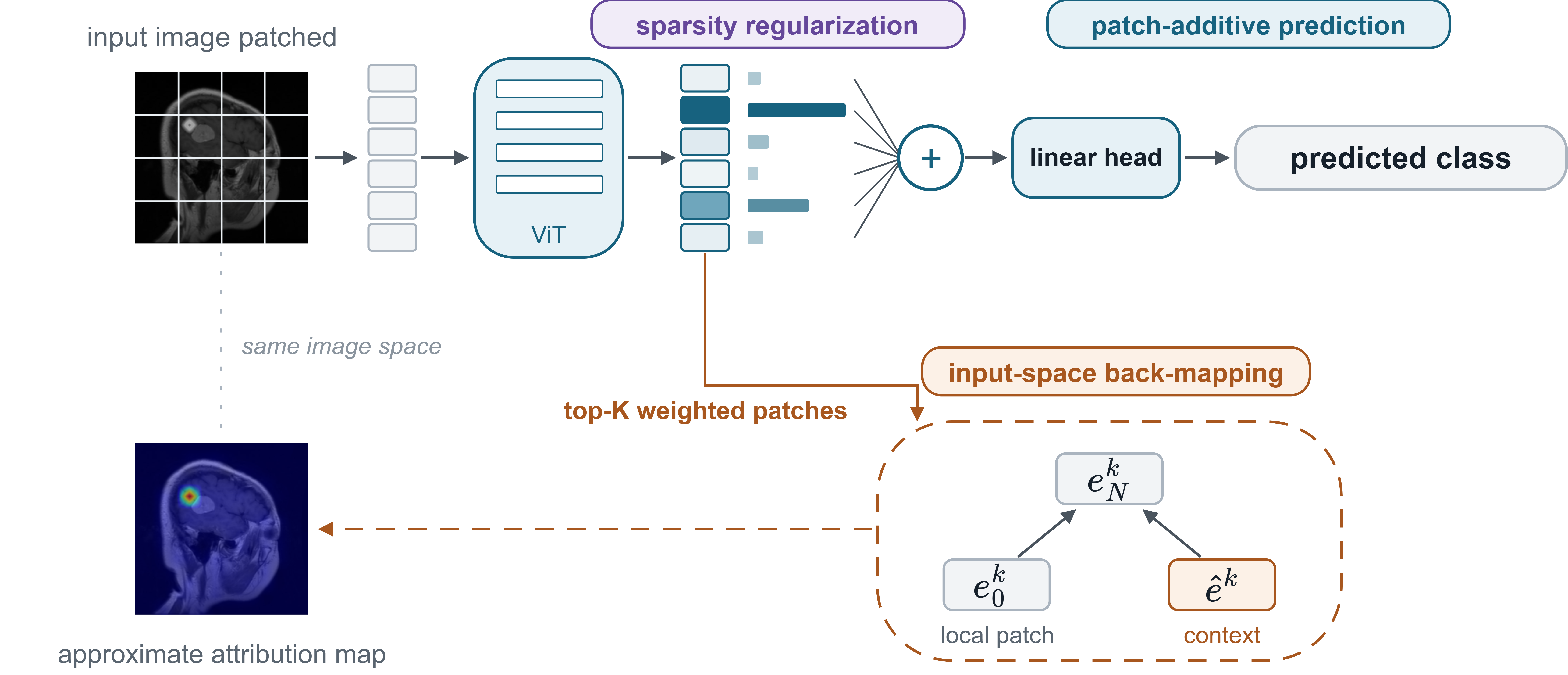}
    \captionsetup{width=\textwidth}
    \caption{\rev{Overview of SAP-ViT.} The input image is divided into patches and processed by the ViT to obtain contextualized patch embeddings. SAP-ViT learns a weight for each output patch, while sparsity regularization encourages the model to focus on a compact set of informative patches. The weighted patch embeddings are additively aggregated and passed to a linear classification head to obtain the predicted class. Because self-attention mixes information across patches, an influential last-layer patch embedding $e_N^k$ may contain both local information from its corresponding input patch embedding $e_0^k$ and contextual information from other image regions, represented by the latent context embedding $\hat{e}^k$. This motivates input-space back-mapping, which uses the pair $(e_0^k,e_N^k)$ to estimate the contextual representation $\hat{e}^k$ through a conditional VAE. The recovered context representation is compared with all input patch embeddings, and the resulting matching maps for the top-$K$ output patches are combined to produce an approximate input-space attribution map.}
    \label{fig:idea}
\end{figure*}

\section{Method}

We propose SAP-ViT, a Self-Attributing Patch Vision Transformer for medical image classification. Figure~\ref{fig:idea} presents an overview of the complete framework and its three main components: patch-additive prediction, sparsity regularization, and input-space back-mapping. The goal of SAP-ViT is to make the image-level prediction decomposable into patch-level contributions while accounting for token mixing in self-attention.

SAP-ViT does not use the final CLS representation for classification. Given an input image, we split it into a sequence of patch tokens and process them using the pretrained Transformer encoder, in which the CLS token remains present during encoding. The final prediction is computed exclusively from the last-layer patch embeddings. This design allows the class score to be decomposed into contributions from individual output patch tokens.

However, because self-attention mixes information across tokens, an output patch embedding should not be interpreted as representing only its corresponding input patch. To address this issue, SAP-ViT introduces an input-space back-mapping module that estimates which input patches contribute to influential output patch embeddings.

\subsection{Preliminaries}

A Vision Transformer (ViT) processes an image as a sequence of patch tokens. Given an input image, we divide it into $M$ non-overlapping patches. Each patch is projected into a feature space and combined with positional encoding, producing the input patch-token sequence
\[
(e^{0}_{0}, \ldots, e^{M-1}_{0})
\]
where $e^{j}_{0}$ denotes the embedding of the $j$-th input patch.

SAP-ViT uses a Transformer encoder with $N$ layers. Each layer takes a sequence of tokens as input and outputs a sequence of updated tokens. Due to the self-attention mechanism, each output token can integrate information from other tokens in the sequence. Therefore, after multiple Transformer layers, each patch embedding becomes context-aware.

Unlike standard ViT classifiers that use the final CLS representation for image-level classification, SAP-ViT computes its prediction exclusively from the final-layer patch embeddings. Suppressing the CLS state for notational simplicity, the patch-token states are written as
\begin{equation}
\bigl[e^{0}_{i}, \ldots, e^{M-1}_{i}\bigr]
= T_i\bigl(\bigl[e^{0}_{i-1}, \ldots, e^{M-1}_{i-1}\bigr]\bigr),
\quad i = 1, \ldots, N
\end{equation}
where $T_i$ denotes the $i$-th Transformer layer.

The final-layer patch embeddings are denoted as
\[
(e^{0}_{N}, \ldots, e^{M-1}_{N})
\]
Each $e^{j}_{N}$ may contain information not only from its corresponding input patch $e^{j}_{0}$, but also from other input patches. This token-mixing property is important for SAP-ViT: it motivates both the patch-additive prediction mechanism and the input-space back-mapping module introduced below.

\subsection{Patch-additive prediction}

Given the final-layer patch embeddings
$(e^{0}_{N}, \ldots, e^{M-1}_{N})$, SAP-ViT computes the
image-level prediction by aggregating patch-level evidence instead of
using the CLS representation for classification. For each output patch
embedding $e^{i}_{N}$, a small multilayer perceptron (MLP) produces an
unnormalized patch score $r_i$:
\begin{equation}
r_i = f_{\omega}(e^{i}_{N}),
\quad i = 0, \ldots, M-1,
\label{eq:patch_score}
\end{equation}
where $f_{\omega}$ denotes the patch-scoring MLP.

The scores are normalized using a softmax function to obtain the
aggregation weights:
\begin{equation}
\alpha_i =
\frac{\exp(r_i)}
{\sum_{j=0}^{M-1}\exp(r_j)},
\quad i = 0, \ldots, M-1.
\label{eq:normalized_weight}
\end{equation}

The final image representation is computed as a weighted sum of the
final-layer patch embeddings:
\begin{equation}
z = \sum_{i=0}^{M-1} \alpha_i e^{i}_{N}.
\label{eq:patch_aggregation}
\end{equation}

The class prediction is obtained using a linear classification head
$h$ parameterized by $W$ and $b$:
\begin{equation}
\ell = h(z) = Wz+b,
\qquad
\hat{c}=\operatorname*{arg\,max}_{c}\ell_c,
\label{eq:prediction}
\end{equation}
where $\ell_c$ denotes the logit for class $c$,
$W_c\in\mathbb{R}^{d}$ denotes the class-specific weight vector such
that $W_c^\top$ is the $c$-th row of $W$, and $b_c$ is the corresponding
bias. Because the classification head is linear, the class logit can be
decomposed as
\begin{equation}
\ell_c
=
b_c+\sum_{i=0}^{M-1}
\underbrace{\alpha_i W_c^{\top}e_N^i}_{a_{i,c}},
\label{eq:class_contribution}
\end{equation}
where $a_{i,c}$ is the contribution of the $i$-th last-layer patch
embedding to class $c$. The normalized aggregation weight $\alpha_i$
determines how strongly the patch embedding enters the image
representation, but it is not alone a class-specific contribution.
The contribution to class $c$ jointly depends on $\alpha_i$, the
contextualized patch embedding $e_N^i$, and the class-specific
classifier parameters $W_c$.

\subsection{Sparsity regularization for selective evidence}

To make the patch-level evidence easier to interpret, we encourage
SAP-ViT to rely on a compact set of informative patch tokens rather
than distributing the aggregation weights diffusely across all
patches. The strength of this sparsity bias is controlled by a
regularization hyperparameter.

We apply an entropy regularization term to the normalized aggregation
weights:
\begin{equation}
\mathcal{L}_{\mathrm{sparse}}
=
H(\alpha)
=
-\sum_{i=0}^{M-1}\alpha_i\log\alpha_i.
\label{eq:sparse_loss}
\end{equation}

The final classifier-training objective is
\begin{equation}
\mathcal{L}
=
\mathcal{L}_{\mathrm{cls}}
+
\lambda_1\mathcal{L}_{\mathrm{sparse}},
\label{eq:classifier_loss}
\end{equation}
where $\mathcal{L}_{\mathrm{cls}}$ is the classification loss and
$\lambda_1$ controls the strength of sparsity regularization.

Minimizing the entropy encourages the aggregation-weight distribution
to become more concentrated, so that a smaller number of output patch
embeddings receive high weights. This improves the readability of the
patch-level evidence while still allowing the model to use multiple
patches when necessary.

\subsection{Input-space back-mapping}
\label{sec:backmapping}

The patch-additive prediction mechanism identifies influential final-layer patch embeddings. However, due to self-attention, an influential output patch embedding should not be interpreted as representing only its corresponding input patch. For the $k$-th input patch embedding $e^{k}_{0}$, the Transformer encoder produces the corresponding final-layer embedding $e^{k}_{N}$. While $e^{k}_{0}$ represents the local input patch, $e^{k}_{N}$ may also contain contextual information from other patches. 
Therefore, when $e^{k}_{N}$ receives a high aggregation weight, we further estimate which input patches are associated with the contextual information contained in this output embedding.

SAP-ViT introduces an input-space back-mapping module for this purpose. Given the pair $(e^{k}_{0}, e^{k}_{N})$, the module estimates a context embedding $\hat{e}^{k}$ that captures information in $e^{k}_{N}$ beyond the local input patch $e^{k}_{0}$. The embeddings $e^{k}_{0}$, $e^{k}_{N}$, and $\hat{e}^{k}$ have the same dimensionality. We model this context estimation using a conditional variational autoencoder (conditional VAE) \cite{vae,c_vae}. 
After training, the projected encoder mean is compared with all input patch embeddings to obtain input-space match scores. A higher match score indicates that an input patch is more aligned with the contextual signal represented in $e^{k}_{N}$.

\subsubsection{Conditional VAE}

We model $e_N^k$ as a function of its local input patch $e_0^k$ and a latent context embedding $\hat{e}^k$. The encoder $q_\phi(\hat{e}^k \mid e_0^k, e_N^k)$ produces a Gaussian posterior, and the decoder $p_\theta(e_N^k \mid \hat{e}^k, e_0^k)$ reconstructs $e_N^k$ from $(\hat{e}^k, e_0^k)$. The training objective $\mathcal{L}_1$ of conditional VAE consists of a reconstruction term and a \rev{Kullback--Leibler (KL)} regularization term:

\begin{equation}
\begin{split}
\mathcal{L}_1
= \mathbb{E}_{k}\Big[
&\mathbb{E}_{q_\phi(\hat{e}^k \mid e_0^{k},e_N^{k})}
\big[-\log p_\theta(e_N^{k}\mid \hat{e}^k,e_0^{k})\big] \\
&+ \beta\,\mathrm{KL}\big(
q_\phi(\hat{e}^k\mid e_0^{k},e_N^{k})
\,\|\,p(\hat{e}^k)\big)
\Big].
\end{split}
\label{eq:cvae_loss}
\end{equation}
where $p(\hat{e}^k)=\mathcal{N}(0,I)$ is the Gaussian prior and $\beta$ controls the strength of the KL regularization.

\subsubsection{Alignment}

The latent embedding $\hat{e}^k$ learned by the conditional VAE is not guaranteed to be directly comparable with the input patch embeddings, because latent variables are only identifiable up to transformations. Consequently, cosine similarity between $\hat{e}^k$ and input tokens $e_0^j$ may not be meaningful without additional constraints. Since our goal is to retrieve input patches aligned with the contextual information in $e_N^k$, we introduce an alignment term that places the estimated context embedding in the input-token space.

Specifically, we take the encoder mean $\mu_\phi(e_0^k, e_N^k)$ and pass it through a lightweight projection head $g_\psi(\cdot)$ that maps the latent representation into the input-token space. The alignment loss penalizes the angular discrepancy between the projected context representation and the corresponding input patch embedding:
\begin{equation}
\mathcal{L}_2
= \mathbb{E}_{k}\!\left[
1-\cos\!\big(g_\psi(\mu_\phi(e_0^{k},e_N^{k})),\,e_0^{k}\big)
\right].
\label{eq:alignment_loss}
\end{equation}

The full objective for the back-mapping module is
\begin{equation}
\mathcal{L}_{\mathrm{map}}
= \mathcal{L}_1 + \lambda_2 \mathcal{L}_2,
\label{eq:mapping_loss}
\end{equation}
where $\lambda_2>0$ controls the strength of the alignment term. The first term encourages the estimated context embedding to reconstruct $e_N^k$ together with the local patch embedding $e_0^k$, while the alignment term provides an inductive bias that makes the estimated context comparable to input patch embeddings.

\subsubsection{Scoring}

\rev{After training the back-mapping module, we compare the projected
encoder mean for the $k$-th output patch with each input patch embedding
$e_0^j$:}
\begin{equation}
s_{k\to j}
=
\cos\!\left(
g_\psi\bigl(\mu_\phi(e_0^k,e_N^k)\bigr),
e_0^j
\right).
\label{eq:matching_score}
\end{equation}

\rev{The score $s_{k\to j}$ measures the alignment between the $j$-th
input patch and the projected latent representation inferred from the
output patch embedding $e_N^k$.}
We use these scores to produce an input-space attribution map. This back-mapping should be interpreted as an approximate attribution mechanism, rather than a guaranteed recovery of the true causal contribution of each input patch.

\rev{For input-space back-mapping, we select output patches using the normalized aggregation weights $\alpha_i$. These weights are input-dependent but class-independent and quantify the relative importance assigned to each patch when constructing the pooled image representation.}
\rev{Let $K$ denote the hyperparameter controlling the number of selected
last-layer output patches, and let}
\begin{equation}
\mathcal{K}
=
\operatorname{TopK}
\left(\{\alpha_i\}_{i=0}^{M-1},K\right),
\qquad
A_j
=
\sum_{k\in\mathcal{K}}s_{k\rightarrow j},
\label{eq:map_aggregation}
\end{equation}
\rev{where $\mathcal{K}$ contains the indices of the $K$ highest-weight
output patches and $A_j$ is the aggregated attribution score of the
$j$-th input patch.}

\rev{Algorithm}~\ref{alg:sapvit}
\rev{summarizes the complete training and inference workflow. SAP-ViT is
trained first, and its parameters are then frozen while the input-space
back-mapping module is trained. During inference, the $K$ highest-weight
output patches are selected, and their input-space matching-score maps are
summed to produce the final attribution map.}

\begin{algorithm}[t]
\caption{\protect\rev{Training and inference workflow of SAP-ViT}}
\label{alg:sapvit}
\begin{algorithmic}[1]

\renewcommand{\algorithmicrequire}{\textbf{Input:}}
\renewcommand{\algorithmicensure}{\textbf{Output:}}

\REQUIRE Training set $\mathcal{D}$, test image $x$, and number of
selected output patches $K$
\ENSURE Predicted class $\hat{c}$ and input-space attribution map $A$

\STATE Train SAP-ViT using patch-additive prediction with sparsity
regularization by minimizing the objective in
Eq.~(\ref{eq:classifier_loss}).

\STATE Freeze the trained SAP-ViT and train the input-space back-mapping
module by minimizing the objective in
Eq.~(\ref{eq:mapping_loss}).

\STATE For $x$, compute the class prediction and normalized patch
weights according to
Eqs.~(\ref{eq:patch_score})--(\ref{eq:prediction}).

\STATE Select the $K$ highest-weight last-layer output patches according
to Eq.~(\ref{eq:map_aggregation}).

\FOR{each selected output patch $k\in\mathcal{K}$}
    \STATE Compute its context embedding and input-patch matching scores
    according to Eq.~(\ref{eq:matching_score}).
\ENDFOR

\STATE Obtain $A$ by summing the matching-score maps according to
Eq.~(\ref{eq:map_aggregation}).

\RETURN $\hat{c}$ and $A$

\end{algorithmic}
\end{algorithm}

\section{Experiment}
\subsection{Setup}

\subsubsection{Datasets}

We evaluate SAP-ViT on four public medical imaging datasets spanning three imaging modalities: brain \rev{magnetic resonance imaging (MRI)}, chest X-ray, and ultrasound. The datasets cover both binary and multi-class classification tasks. The brain tumor MRI dataset \cite{dataset_brain} is a four-class classification task including glioma, meningioma, no tumor, and pituitary tumor. The COVID chest X-ray dataset \cite{dataset_COVID_1,dataset_COVID_2} is a four-class task including COVID, lung opacity, normal, and viral pneumonia. The pneumonia chest X-ray dataset \cite{dataset_chest} is a binary classification task including normal and pneumonia. The breast ultrasound dataset \cite{dataset_breast} is a three-class task including normal, benign, and malignant.

For each dataset, we resize all images to a fixed resolution and apply the same preprocessing pipeline across SAP-ViT and all baselines. We split each dataset into training, validation, and test sets, and report results on the test set. Table~\ref{tab:dataset_statistics} summarizes the number of classes and the number of images used for each dataset.




\subsubsection{Predictive and explainability baselines}

\rev{For predictive-performance comparison, we include two model-level
baselines: a standard CLS-based vanilla ViT and IA-ViT}
\cite{ref_ia_1}\rev{. The vanilla ViT uses the same Transformer
configuration and training protocol as SAP-ViT but retains the CLS token
and the standard CLS-based classification head. IA-ViT jointly trains an
interpreter and a predictor. SAP-ViT, vanilla ViT, and IA-ViT are compared
in terms of classification performance.}

\rev{For explanation evaluation, we compare SAP-ViT and IA-ViT with four
post-hoc explanation methods: Grad-CAM}\cite{ref_grad_cam}\rev{,
attention rollout}\cite{ref_att_rollout}\rev{, LIME}\cite{ref_lime}
\rev{, and SHAP}\cite{ref_shap}\rev{. All four post-hoc methods are
applied to the same trained vanilla ViT described above. They therefore
do not constitute separate predictive models and are used solely for
explanation evaluation. The classification performance of their shared
vanilla ViT backbone is reported only once.}

\rev{To support reproducibility}, the complete implementation code and all
hyperparameter configurations for SAP-ViT are publicly available in our
\href{https://github.com/Wienhannover/SAP-ViT}{GitHub repository}.

\begin{table}[t]
\centering
\caption{Dataset statistics.}
\label{tab:dataset_statistics}
\begin{tabular}{lccc}
\hline
Dataset & Modality & \# Classes & \# Images \\
\hline
Brain tumor MRI & MRI & 4 & 7023 \\
COVID chest X-ray & Chest X-ray & 4 & 21165 \\
Pneumonia chest X-ray & Chest X-ray & 2 & 5856 \\
Breast ultrasound & Ultrasound & 3 & 780 \\
\hline
\end{tabular}
\end{table}

\subsection{Results}

\subsubsection{Model performance}

\begin{table*}[ht]
    \caption{Model performance measured by classification accuracy.}
    \centering
    \begin{adjustbox}{max width=\textwidth}
    \begin{tabular}{@{}lcccc@{}}
        \toprule
         & \multicolumn{4}{c}{} \\
        
              & \makecell[c]{Brain tumor\\ MRI} & \makecell[c]{COVID\\chest X-ray} & \makecell[c]{Pneumonia\\chest X-ray} & \makecell[c]{Breast\\ultrasound} \\
        \midrule
        SAP-ViT & $0.993 \pm 0.003$ & $0.955 \pm 0.003$ & $0.909 \pm 0.019$	 & $0.878 \pm 0.019$ \\
        IA-ViT     & $0.905 \pm 0.020$ & $0.658 \pm 0.066$ & $0.713 \pm 0.067$ & $0.559 \pm 0.006$ \\
        \midrule
        \makecell[l]{Vanilla ViT\\(post-hoc backbone)}
                   & $0.973 \pm 0.008$ & $0.929 \pm 0.006$ & $0.888 \pm 0.025$ & $0.780 \pm 0.074$ \\
        \bottomrule
    \end{tabular}
    \end{adjustbox}
    \label{table_2}
\end{table*}

Table~\ref{table_2} \rev{reports classification accuracy for the models
evaluated under the same experimental protocol. Among these implemented
models, SAP-ViT achieves the highest mean accuracy on all four datasets,
including brain tumor MRI, COVID chest X-ray, pneumonia chest X-ray, and
breast ultrasound.} Additional classification metrics are reported in Appendix~\ref{app:classification_metrics}. As shown in Table~\ref{tab:classification_more_metrics}, SAP-ViT also achieves the highest mean macro-averaged precision, sensitivity, and F1 score among the implemented models on all four datasets.

Compared with IA-ViT, SAP-ViT obtains substantially higher accuracy on all four datasets. This indicates that the proposed patch-level self-attributing design does not compromise predictive performance compared with the interpretable-by-design baseline. Compared with the vanilla ViT model used for the post-hoc explanation baselines, SAP-ViT also achieves higher accuracy across all datasets, showing that the proposed architecture remains competitive with standard ViT-based classification while providing intrinsic patch-level attribution.



\rev{To contextualize SAP-ViT's predictive performance},
Appendix~\ref{app:published_comparison} compares SAP-ViT with
representative recent studies~\cite{ref_disci,ref_neuronet,ref_shahin}
evaluated on the same brain tumor MRI dataset. Despite
differences in data splits, preprocessing, and training
protocols, SAP-ViT achieves classification performance comparable to
that reported by these studies. Its principal advantage lies in combining
competitive predictive performance with an intrinsic patch-attribution mechanism, rather than universally outperforming existing classifiers in accuracy. We next evaluate the faithfulness of
the resulting attribution maps.

\subsubsection{Explanation evaluation}

We evaluate explanation faithfulness using insertion ($I$), deletion ($D$), and their difference ($I-D$). A higher insertion score indicates that the most important regions identified by the explanation are sufficient to recover the model prediction when progressively inserted into a masked image. A lower deletion score indicates that removing the most important regions more strongly suppresses the prediction. Therefore, a larger $I-D$ gap reflects explanations that are both more sufficient under insertion and more necessary under deletion.

\rev{These perturbation-based metrics evaluate explanation faithfulness
with respect to the model's prediction under the selected insertion and
deletion protocol. They do not establish that the highlighted regions are
clinically correct, causally related to the disease, or consistent with
expert interpretation. Establishing clinical validity would require
expert assessment or ground-truth lesion annotations, which are beyond the
scope of this study.}

\begin{table*}[ht]
    \caption{Explanation evaluation. $I$ denotes the insertion score, $D$ denotes the deletion score, and $I-D$ represents the gap between insertion and deletion. The last column reports the mean $I-D$ across all datasets.}
    \centering
    \begin{adjustbox}{max width=\textwidth}
    \begin{tabular}{@{}lcccccc@{}}
        \toprule
         & & \makecell[c]{Brain tumor\\ MRI} 
           & \makecell[c]{COVID\\chest X-ray} 
           & \makecell[c]{Pneumonia\\chest X-ray} 
           & \makecell[c]{Breast\\ultrasound}
           & \makecell[c]{Mean\\$I-D$} \\ 
        \midrule
        
        \multirow{3}{*}{SAP-ViT} 
        & $D$   & $0.400 \pm 0.020$  & $0.449 \pm 0.025$ & $0.659 \pm 0.013$ & $0.522 \pm 0.024$ & -- \\ 
        & $I$   & $0.733 \pm 0.007$  & $0.677 \pm 0.033$ & $0.789 \pm 0.025$ & $0.693 \pm 0.022$ & -- \\   
        & $I-D$ & $\textbf{0.333} \pm \textbf{0.022}$  
                 & $\textbf{0.228} \pm \textbf{0.035}$ 
                 & $0.130 \pm 0.024$ 
                 & $\textbf{0.171} \pm \textbf{0.012}$ 
                 & $\textbf{0.216}$ \\
        \midrule

        \multirow{3}{*}{IA-ViT}
        & $D$   & $0.423 \pm 0.015$ & $0.425 \pm 0.043$ & $0.580 \pm 0.056$ & $0.408 \pm 0.010$ & -- \\
        & $I$   & $0.416 \pm 0.020$ & $0.439 \pm 0.031$ & $0.577 \pm 0.023$ & $0.412 \pm 0.025$ & -- \\
        & $I-D$ & $-0.007 \pm 0.013$ & $0.014 \pm 0.031$ & $-0.003 \pm 0.047$ & $0.004 \pm 0.027$ & $0.002$ \\
        \midrule

        \multirow{3}{*}{LIME}
        & $D$   & $0.380 \pm 0.015$ & $0.363 \pm 0.042$ & $0.608 \pm 0.040$ & $0.502 \pm 0.059$ & -- \\
        & $I$   & $0.518 \pm 0.041$ & $0.484 \pm 0.067$ & $0.856 \pm 0.038$ & $0.568 \pm 0.020$ & -- \\
        & $I-D$ & $0.138 \pm 0.030$ & $0.121 \pm 0.047$ 
                 & $\textbf{0.248} \pm \textbf{0.066}$ 
                 & $0.066 \pm 0.048$ 
                 & $0.143$ \\
        \midrule

        \multirow{3}{*}{SHAP}
        & $D$   & $0.401 \pm 0.024$ & $0.413 \pm 0.051$ & $0.714 \pm 0.008$ & $0.581 \pm 0.015$ & -- \\
        & $I$   & $0.490 \pm 0.031$ & $0.481 \pm 0.063$ & $0.791 \pm 0.030$ & $0.568 \pm 0.034$ & -- \\
        & $I-D$ & $0.089 \pm 0.010$ & $0.068 \pm 0.021$ & $0.076 \pm 0.028$ & $-0.013 \pm 0.024$ & $0.055$ \\
        \midrule

        \multirow{3}{*}{Grad-CAM}
        & $D$   & $0.402 \pm 0.023$ & $0.461 \pm 0.060$ & $0.694 \pm 0.033$ & $0.506 \pm 0.062$ & -- \\
        & $I$   & $0.519 \pm 0.053$ & $0.515 \pm 0.031$ & $0.881 \pm 0.036$ & $0.669 \pm 0.025$ & -- \\
        & $I-D$ & $0.117 \pm 0.039$ & $0.055 \pm 0.049$ & $0.187 \pm 0.049$ & $0.163 \pm 0.051$ & $0.131$ \\
        \midrule

        \multirow{3}{*}{Rollout}
        & $D$   & $0.371 \pm 0.018$ & $0.452 \pm 0.048$ & $0.689 \pm 0.019$ & $0.533 \pm 0.034$ & -- \\
        & $I$   & $0.554 \pm 0.035$ & $0.595 \pm 0.053$ & $0.825 \pm 0.010$ & $0.599 \pm 0.036$ & -- \\
        & $I-D$ & $0.183 \pm 0.022$ & $0.143 \pm 0.044$ & $0.136 \pm 0.017$ & $0.067 \pm 0.014$ & $0.132$ \\
        \bottomrule
    \end{tabular}
    \end{adjustbox}
    \label{table_1}
\end{table*}

As shown in Table~\ref{table_1}, SAP-ViT achieves the highest mean $I-D$ across all datasets, indicating the most consistent overall explanation quality among the evaluated methods. In addition, SAP-ViT obtains the best $I-D$ score on three of four datasets: brain tumor MRI, COVID chest X-ray, and breast ultrasound.

Compared with the interpretable-by-design baseline IA-ViT, SAP-ViT produces substantially larger $I-D$ gaps on all datasets. IA-ViT obtains near-zero or negative $I-D$ values, suggesting that its explanation maps do not consistently identify regions that are both sufficient and necessary for the prediction. In contrast, SAP-ViT explicitly decomposes the prediction into sparse patch-level contributions, leading to more discriminative attribution maps.

Compared with post-hoc baselines, SAP-ViT also shows stronger overall performance. It outperforms LIME, SHAP, Grad-CAM, and Rollout on brain tumor MRI and COVID chest X-ray, and achieves the best result on breast ultrasound. These results suggest that integrating attribution into the prediction mechanism can improve explanation faithfulness while maintaining competitive performance.

 SAP-ViT does not achieve the best $I-D$ score on the pneumonia chest X-ray dataset. This suggests that, although SAP-ViT provides the strongest overall performance across datasets, its sparse patch-level attribution mechanism may not be equally advantageous in all imaging settings. Pneumonia chest X-ray may represent a case where discriminative evidence is more spatially distributed, making it less aligned with explanations that emphasize a small number of highly influential patches.
 
 To further examine whether this dataset-dependent behavior is related to image characteristics, we conducted a model-independent patch-level radiomics-style concentration analysis. For each dataset, we sampled up to 500 images, computed simple patch-level texture features, summarized their spatial concentration per image, and averaged the resulting metrics at the dataset level. We report normalized entropy and Gini coefficient as complementary measures. Higher normalized entropy indicates a more spatially diffuse patch-level texture distribution, whereas higher Gini coefficient indicates stronger spatial concentration.

 \begin{figure}[!t]
    \centering
    \includegraphics[width=\columnwidth]{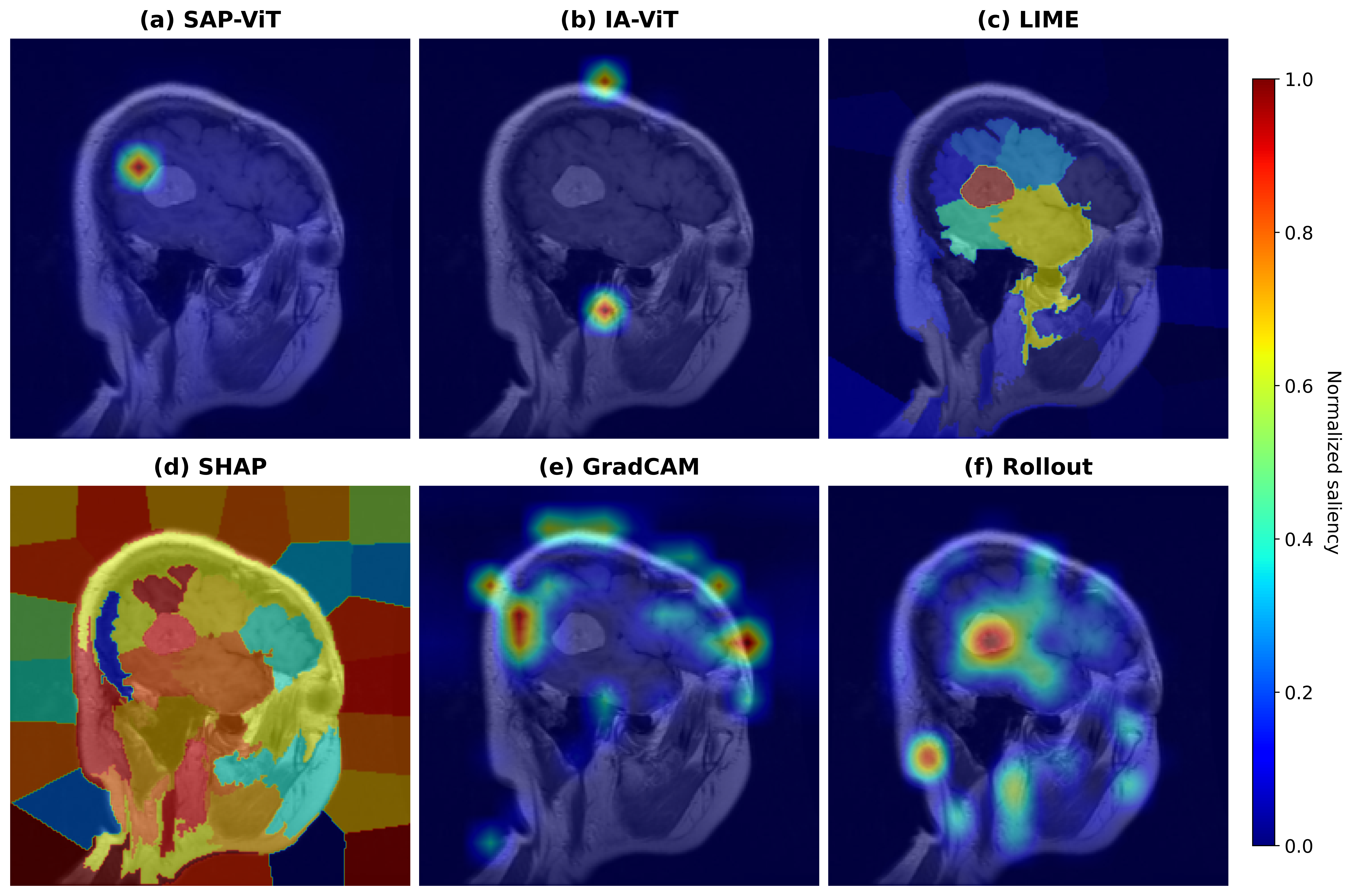}
    \caption{Qualitative attribution comparison on a Brain Tumor MRI example. The ground-truth label is meningioma. SAP-ViT produces a compact attribution map concentrated around the tumor region, while the baseline methods generate either less localized or more diffuse saliency patterns across surrounding anatomical structures and background regions.}
    \label{fig:qualitative}
\end{figure}

\begin{table}[ht]
    \caption{Model-independent patch-level radiomics-style concentration analysis. 
    Entropy$_{\mathrm{norm}}$ measures the spatial spread of patch-level texture responses, 
    where higher values indicate more diffuse responses. Gini measures concentration, 
    where higher values indicate stronger spatial concentration.}
    \centering
    \begin{tabular}{lcc}
        \toprule
        Dataset & Entropy$_{\mathrm{norm}}$ & Gini \\
        \midrule
        Brain tumor MRI        & $0.901$ & $0.476$ \\
        COVID chest X-ray      & $0.949$ & $0.333$ \\
        Pneumonia chest X-ray  & $0.953$ & $0.322$ \\
        Breast ultrasound      & $0.705$ & $0.488$ \\
        \bottomrule
    \end{tabular}
    \label{tab:radiomics_concentration}
\end{table}

As shown in Table~\ref{tab:radiomics_concentration}, breast
ultrasound has the lowest normalized entropy and the highest Gini
coefficient, followed by brain tumor MRI, indicating that their
patch-level texture responses are relatively spatially concentrated.
SAP-ViT achieves the best $I-D$ scores on both datasets, which is
consistent with its main advantage: sparse patch aggregation can isolate
a compact set of influential output patches, while input-space
back-mapping approximately relates their contextualized information to
the corresponding input regions.

\rev{In contrast}, both chest X-ray datasets exhibit higher normalized entropy
and lower Gini values, indicating more spatially diffuse texture
responses. Pneumonia chest X-ray has the highest normalized entropy and
the lowest Gini coefficient, providing one possible explanation for why
SAP-ViT does not achieve the best $I-D$ score on this dataset: its
sparsity bias may under-represent information distributed across broad
image regions. Nevertheless, SAP-ViT achieves the best $I-D$ score on
the COVID chest X-ray dataset despite its relatively diffuse texture
responses. This result indicates that SAP-ViT is not restricted to
strongly localized image patterns and can still identify compact,
prediction-relevant evidence within globally diffuse images.

The radiomics-style analysis is complementary rather than conclusive,
because these model-independent texture statistics do not directly
measure class-discriminative evidence. Overall, the results suggest that
SAP-ViT is particularly advantageous when informative patterns can be
represented by a compact set of patches, while remaining effective
across datasets with substantially different spatial characteristics.

\subsubsection{Ablation study}

\begin{table}[t]
\caption{Ablation study of SAP-ViT on the Brain Tumor MRI dataset.}
\centering
\begin{adjustbox}{max width=\columnwidth}
\begin{tabular}{lccc}
\toprule
Variant & Deletion $\downarrow$ & Insertion $\uparrow$ & $I-D$ $\uparrow$ \\
\midrule
SAP-ViT w/o sparsity
& $0.553 \pm 0.052$
& $0.493 \pm 0.023$
& $-0.060 \pm 0.058$ \\

SAP-ViT w/o alignment
& $0.490 \pm 0.079$
& $0.577 \pm 0.081$
& $0.087 \pm 0.158$ \\

Full SAP-ViT
& $0.400 \pm 0.020$
& $0.733 \pm 0.007$
& $0.333 \pm 0.022$ \\
\bottomrule
\end{tabular}
\end{adjustbox}
\label{tab:ablation}
\end{table}

Table~\ref{tab:ablation} reports the ablation study on the Brain Tumor MRI dataset. Removing the sparsity regularization leads to a substantial drop in explanation quality, with the $I-D$ score becoming negative. This indicates that without sparsity, the model no longer identifies compact and discriminative evidence regions, and the resulting attribution becomes less faithful under insertion and deletion evaluation.

Removing the alignment term also degrades performance, although the drop is less severe than removing sparsity. This suggests that the alignment component is important for making the recovered context embedding comparable to the input patch space, thereby improving the quality of input-space back-mapping. The full SAP-ViT achieves the best deletion, insertion, and $I-D$ scores, showing that both sparsity and alignment contribute to the final explanation quality.

\rev{To examine whether these findings generalize across datasets}, we further conduct the same ablation study on COVID chest X-ray, pneumonia chest X-ray, and breast ultrasound. The corresponding results are reported in Appendix~\ref{app:additional_ablation}, specifically in Table~\ref{tab:additional_ablation}. The results show a consistent trend across all three datasets: removing sparsity causes the largest degradation, while removing alignment also reduces explanation quality compared with the full SAP-ViT.

\subsubsection{Qualitative attribution analysis}

Figure~\ref{fig:qualitative} shows a qualitative comparison on the Brain Tumor MRI dataset. The example contains a visible tumor region in the left intracranial area. SAP-ViT produces a compact and highly localized attribution map, with the strongest response concentrated around the tumor region and limited activation elsewhere. This suggests that SAP-ViT identifies a small set of discriminative patches that are spatially aligned with the lesion.

In contrast, IA-ViT also produces sparse high-saliency regions, but these regions are less aligned with the tumor and appear around regions outside the visually apparent tumor area, such as the skull boundary and lower head region. 
LIME highlights a broader superpixel region that partially overlaps with the tumor, but the attribution also extends to surrounding brain areas, making the localization less precise. SHAP produces fragmented and widespread superpixel regions across the image, including large areas outside the lesion and background, which makes the explanation harder to interpret as localized medical evidence.

Grad-CAM and Rollout generate smoother heatmaps, but their saliency is more diffuse than SAP-ViT. Grad-CAM assigns high responses to multiple regions around the skull boundary and image periphery, while Rollout spreads attribution over a larger part of the brain and nearby anatomical structures. Although these methods capture some relevant image regions, their explanations are less spatially specific. Overall, this qualitative example shows that SAP-ViT provides more localized and visually interpretable attribution than the evaluated baselines.

\rev{Additional qualitative comparisons for other datasets} are provided in
Appendix~\ref{app:additional_qualitative}. Specifically,
Figs.~\ref{fig:qualitative_covid}, \ref{fig:qualitative_pneumonia}, and
\ref{fig:qualitative_breast} present the corresponding attribution
comparisons. Across these examples, SAP-ViT generally produces more compact
attribution patterns, whereas the baseline methods tend to generate less
localized or more diffuse responses. 

\subsection{Limitations and applicability}

\rev{SAP-ViT is most applicable when the discriminative evidence can be
represented by a relatively compact set of image regions. Its entropy
regularization introduces a sparsity bias rather than an assumption that is
universally valid. When evidence is spatially distributed or multifocal,
strong sparsity may suppress complementary regions and reduce explanation
faithfulness, as suggested by the pneumonia chest X-ray results. The
regularization strength therefore represents a dataset-dependent trade-off
between attribution concentration and coverage of distributed evidence.
Furthermore, input-space back-mapping provides approximate associations
rather than causal attribution, while insertion and deletion evaluate
faithfulness to model behavior rather than clinical correctness. Finally,
the present evaluation is limited to four public datasets spanning MRI,
chest X-ray, and ultrasound; further evaluation is required before
generalizing the findings to other medical imaging settings.}

\section{Conclusion}
In this paper, we proposed SAP-ViT, a self-attributing Vision Transformer for medical image classification. Rather than relying on a global CLS token for prediction, SAP-ViT directly aggregates last-layer patch embeddings, enabling class predictions to be decomposed into patch-level contributions. To further improve interpretability, SAP-ViT incorporates a sparsity regularization term that encourages the model to focus on a compact set of informative patches, together with an input-space back-mapping mechanism that mitigates the ambiguity introduced by token mixing in self-attention.

Experiments on multiple medical imaging datasets demonstrate that SAP-ViT maintains strong classification performance while improving the interpretability of ViT predictions. Overall, SAP-ViT provides a favorable trade-off between predictive accuracy and interpretability, making it a promising framework for transparent ViT-based medical image classification.

\FloatBarrier
\appendices

\section{Additional Ablation Results}
\label{app:additional_ablation}

Table~\ref{tab:additional_ablation} reports the ablation results on the three
additional datasets. Consistent with the brain-tumor MRI results presented
in the main text, the complete SAP-ViT outperforms the variants without
sparsity regularization or alignment.

\begin{table}[!htbp]
\centering
\caption{Additional ablation studies of SAP-ViT on the COVID chest X-ray, Pneumonia chest X-ray, and Breast Ultrasound datasets.}
\label{tab:additional_ablation}
\begin{adjustbox}{max width=\columnwidth}
\begin{tabular}{lccc}
\toprule
Variant & Deletion $\downarrow$ & Insertion $\uparrow$ & $I-D$ $\uparrow$ \\
\midrule
\multicolumn{4}{l}{\textbf{COVID chest X-ray}} \\
SAP-ViT w/o sparsity
& $0.559 \pm 0.056$
& $0.598 \pm 0.084$
& $0.039 \pm 0.113$ \\
SAP-ViT w/o alignment
& $0.454 \pm 0.024$
& $0.656 \pm 0.046$
& $0.202 \pm 0.044$ \\
Full SAP-ViT
& $0.449 \pm 0.025$
& $0.677 \pm 0.033$
& $0.228 \pm 0.035$ \\
\midrule

\multicolumn{4}{l}{\textbf{Pneumonia chest X-ray}} \\
SAP-ViT w/o sparsity
& $0.716 \pm 0.024$
& $0.770 \pm 0.022$
& $0.054 \pm 0.022$ \\
SAP-ViT w/o alignment
& $0.668 \pm 0.024$
& $0.774 \pm 0.038$
& $0.106 \pm 0.044$ \\
Full SAP-ViT
& $0.659 \pm 0.013$
& $0.789 \pm 0.025$
& $0.130 \pm 0.024$ \\
\midrule

\multicolumn{4}{l}{\textbf{Breast ultrasound}} \\
SAP-ViT w/o sparsity
& $0.554 \pm 0.015$
& $0.656 \pm 0.029$
& $0.103 \pm 0.031$ \\
SAP-ViT w/o alignment
& $0.528 \pm 0.037$
& $0.689 \pm 0.013$
& $0.161 \pm 0.025$ \\
Full SAP-ViT
& $0.522 \pm 0.024$
& $0.693 \pm 0.022$
& $0.171 \pm 0.012$ \\
\bottomrule
\end{tabular}
\end{adjustbox}
\end{table}

\section{Additional Qualitative Attribution Examples}
\label{app:additional_qualitative}

\begin{figure}[!htbp]
    \centering
    \includegraphics[width=\columnwidth]
    {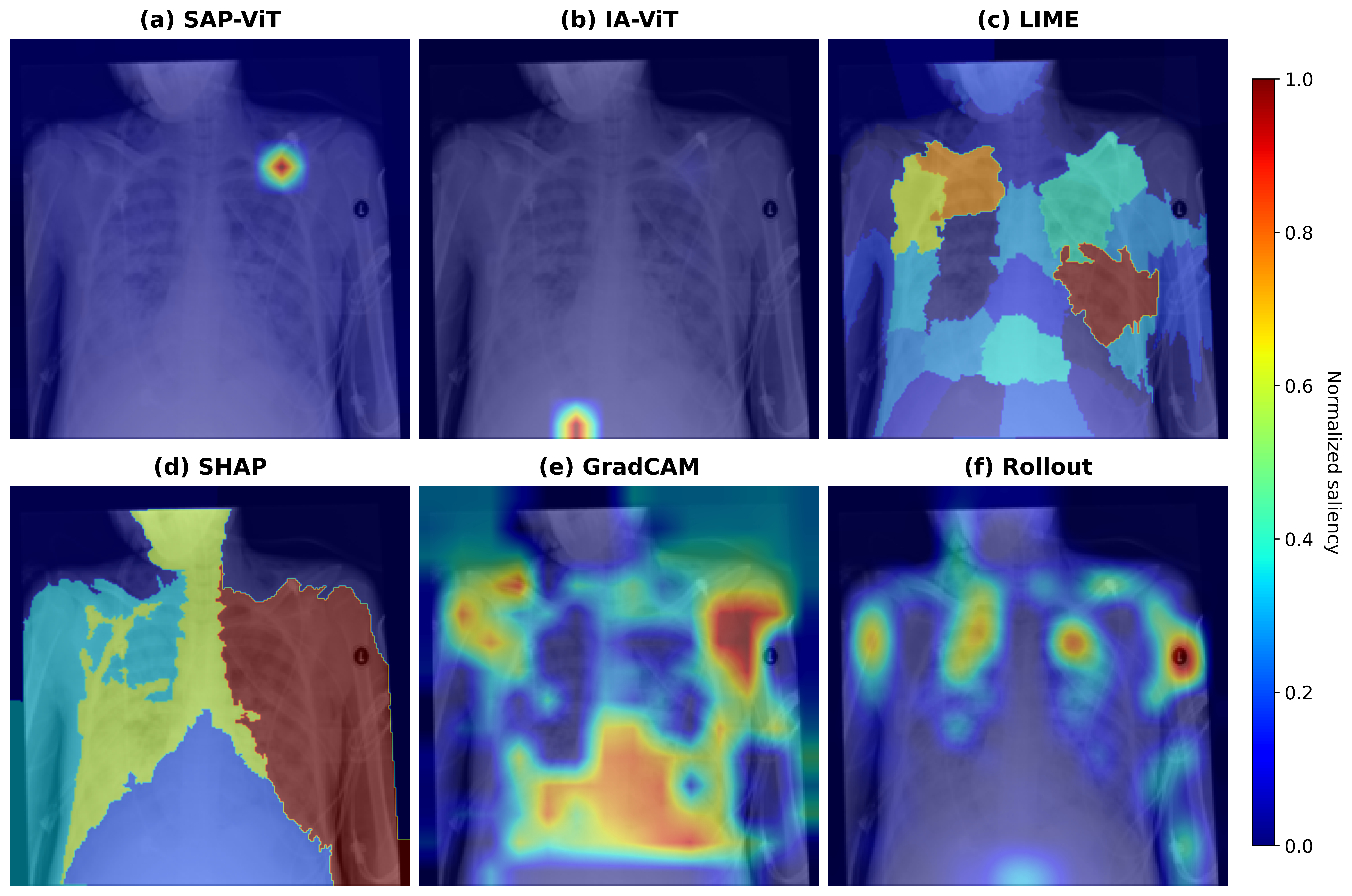}
    \caption{Qualitative attribution comparison on a chest radiograph
    from the COVID chest X-ray dataset with the ground-truth label lung
    opacity. SAP-ViT produces a compact attribution map concentrated in an
    upper-lung region exhibiting visible opacity, whereas the baseline
    methods generate less localized or more diffuse saliency patterns
    across the lungs, surrounding anatomical structures, and background.}
    \label{fig:qualitative_covid}
\end{figure}

\begin{figure}[!htbp]
    \centering
    \includegraphics[width=\columnwidth]
    {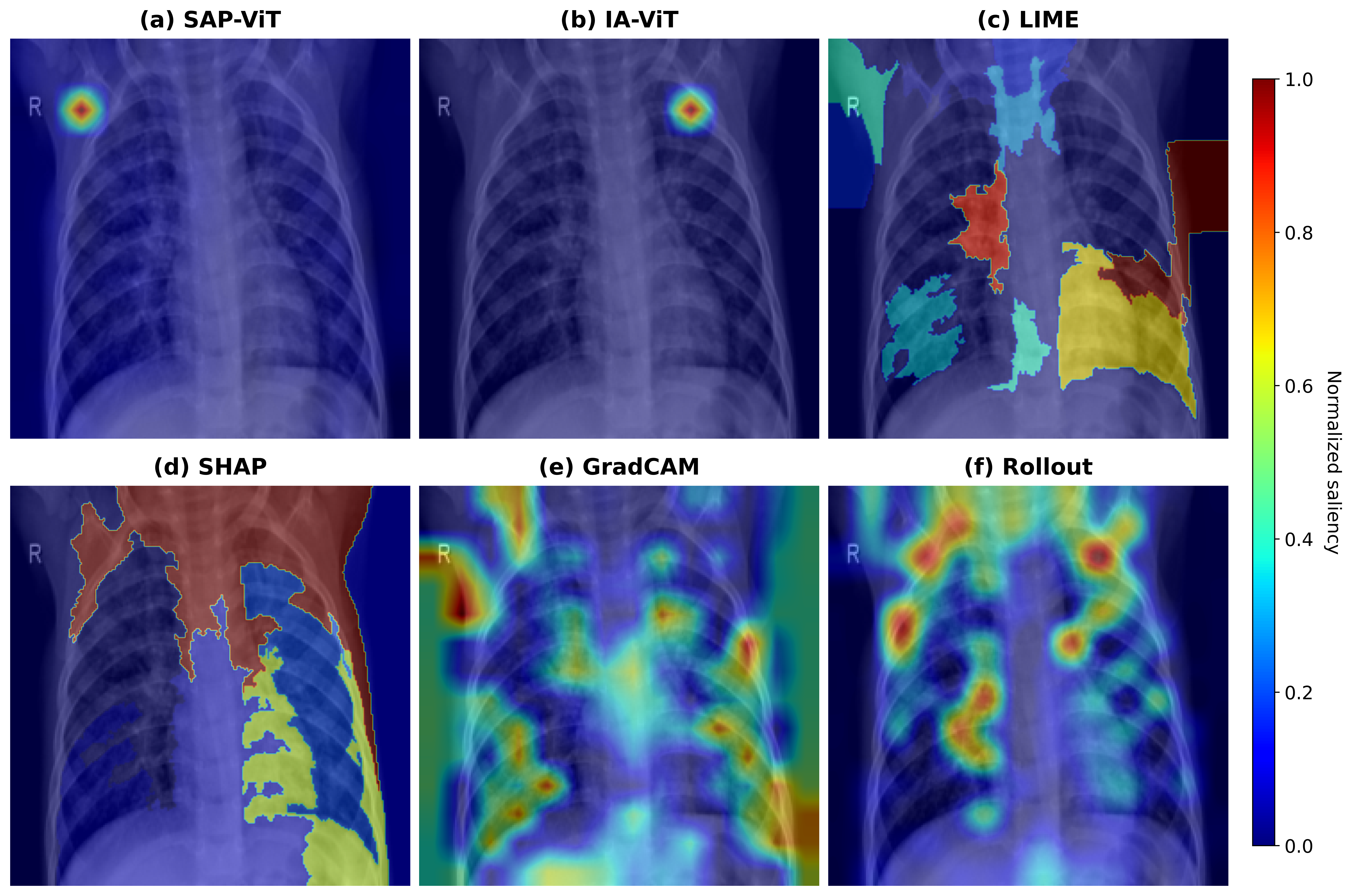}
    \caption{Qualitative attribution comparison on an example from the
    pneumonia chest X-ray dataset with the ground-truth label pneumonia.
    SAP-ViT produces a compact attribution map concentrated in an upper-lung region exhibiting visible opacity, whereas the baseline
    methods generate less localized or more diffuse saliency patterns
    across the lungs, surrounding anatomical structures, and background.}
    \label{fig:qualitative_pneumonia}
\end{figure}

\begin{figure}[!htbp]
    \centering
    \includegraphics[width=\columnwidth]
    {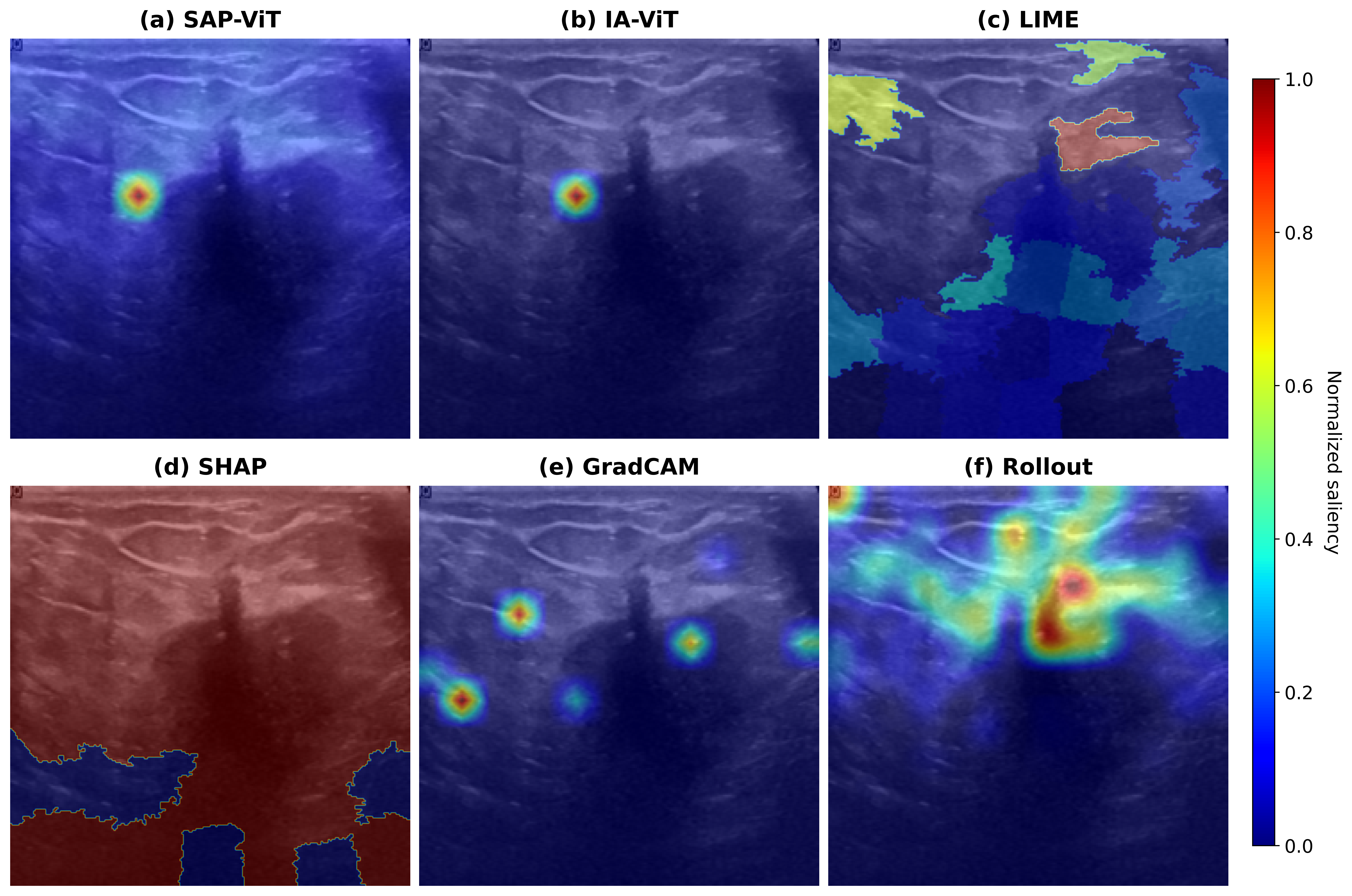}
    \caption{Qualitative attribution comparison on a breast ultrasound example. The ground-truth label is malignant. SAP-ViT produces a compact attribution map concentrated around the lesion, with its strongest hotspot overlapping the boundary of the supplied tumor mask. In contrast, the baseline methods generate either less localized or more diffuse saliency patterns across the surrounding tissue and background regions.}
    \label{fig:qualitative_breast}
    
\end{figure}

Figures~\ref{fig:qualitative_covid},
\ref{fig:qualitative_pneumonia}, and
\ref{fig:qualitative_breast} provide qualitative attribution comparisons
on the three additional datasets. Within each comparison, all methods are
applied to the same input image. These examples illustrate the spatial
characteristics of the attribution maps and are intended to complement the dataset-level quantitative evaluation.

\section{Comparison with Published Classification Results}
\label{app:published_comparison}

Table~\ref{tab:published_brain_results} compares SAP-ViT with
representative recent classification results reported on the same brain tumor MRI dataset. Because the studies use different
experimental protocols, the values provide literature context rather than
a controlled head-to-head comparison.

\begin{table}[!htbp]
\centering
\caption{Comparison with representative published classification results.}
\label{tab:published_brain_results}
\begin{tabular}{lcc}
\toprule
Study & Model & Accuracy (\%) \\
\midrule
Disci et al.~\cite{ref_disci} & Xception & 98.73 \\
Haque et al.~\cite{ref_neuronet} & NeuroNet19 & 99.30 \\
Shahin~\cite{ref_shahin} & Fine-tuned ResNet34 & 99.66 \\
SAP-ViT (ours) & SAP-ViT & $99.30 \pm 0.30$ \\
\bottomrule
\end{tabular}
\end{table}

\section{Additional Classification Metrics}
\label{app:classification_metrics}

Table~\ref{tab:classification_more_metrics} supplements the accuracy results in Table~\ref{table_2} with macro-averaged precision, sensitivity, and F1 score. Macro-averaging assigns equal importance to every class, providing a class-balanced assessment of predictive performance.

\begin{table}[!htbp]
\centering
\caption{Additional classification metrics for models evaluated under the
same experimental protocol.}
\label{tab:classification_more_metrics}
\begin{tabular}{lccc}
\toprule
Model & Precision & Sensitivity & F1 \\
\midrule
\multicolumn{4}{l}{\textbf{Brain tumor MRI}} \\
SAP-ViT & $0.992 \pm 0.003$ & $0.992 \pm 0.003$ & $0.992 \pm 0.003$ \\
IA-ViT & $0.903 \pm 0.020$ & $0.900 \pm 0.022$ & $0.901 \pm 0.021$ \\
Vanilla ViT & $0.972 \pm 0.008$ & $0.971 \pm 0.009$ & $0.971 \pm 0.009$ \\
\midrule

\multicolumn{4}{l}{\textbf{COVID chest X-ray}} \\
SAP-ViT & $0.965 \pm 0.004$ & $0.959 \pm 0.004$ & $0.962 \pm 0.002$ \\
IA-ViT & $0.560 \pm 0.193$ & $0.495 \pm 0.151$ & $0.462 \pm 0.176$ \\
Vanilla ViT & $0.944 \pm 0.003$ & $0.929 \pm 0.009$ & $0.935 \pm 0.006$ \\
\midrule

\multicolumn{4}{l}{\textbf{Pneumonia chest X-ray}} \\
SAP-ViT & $0.934 \pm 0.011$ & $0.880 \pm 0.026$ & $0.897 \pm 0.024$ \\
IA-ViT & $0.672 \pm 0.208$ & $0.644 \pm 0.108$ & $0.616 \pm 0.159$ \\
Vanilla ViT & $0.920 \pm 0.013$ & $0.853 \pm 0.034$ & $0.872 \pm 0.032$ \\
\midrule

\multicolumn{4}{l}{\textbf{Breast ultrasound}} \\
SAP-ViT & $0.871 \pm 0.018$ & $0.865 \pm 0.028$ & $0.866 \pm 0.019$ \\
IA-ViT & $0.215 \pm 0.065$ & $0.341 \pm 0.018$ & $0.256 \pm 0.040$ \\
Vanilla ViT & $0.805 \pm 0.051$ & $0.721 \pm 0.127$ & $0.733 \pm 0.111$ \\
\bottomrule
\end{tabular}
\end{table}

\FloatBarrier

\newpage
\begin{IEEEbiography}[{\includegraphics[width=1in,height=1.25in,clip,keepaspectratio]{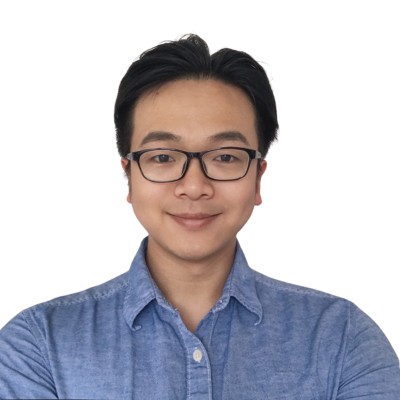}}]{Wei Zhang} received the B.Sc. degree in Computer Science from Xidian University, China, and the M.Sc. degree in Computer Science from Leibniz University Hannover, Germany. He is currently pursuing the Ph.D. degree with the Faculty of Electrical Engineering and Computer Science at Leibniz University Hannover. He is working as a Research Assistant with the L3S Research Center, Hannover. He is involved in the CAIMed project (Lower Saxony Center for AI and Causal Methods in Medicine), where he focuses on uncovering the underlying biological mechanisms of post-acute infection symptoms, specifically long COVID. His research interests include interpretability of neural networks, particularly in the areas of graph neural networks and vision transformers; the integration of causality and reinforcement learning; and foundation models in biomedicine, with a focus on predicting gene expression profiles under gene perturbations.
\end{IEEEbiography}

\begin{IEEEbiography}[{\includegraphics[width=1in,height=1.25in,clip,keepaspectratio]{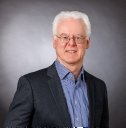}}]{Wolfgang Nejdl}  was born in 1960.
He received the Dipl.Ing. and Dr.Techn. degrees
in computer science from Vienna University
of Technology, in 1984 and 1988, respectively.
He has been a Full Professor of computer
science with Leibniz University Hannover, since
1995. He was an Assistant Professor in Vienna,
from 1988 to 1992, and an Associate Professor
at RWTH Aachen University, from 1992 to 1995.
He was a Visiting Researcher/a Professor with
Xerox PARC, Stanford University, the University of Illinois at UrbanaChampaign, EPFL Lausanne, PUC Rio, Trento, and Politecnico di
Milano. He heads the L3S Research Center, www.L3S.de, and the
Data Science Institute/Knowledge-Based Systems at Leibniz University.
From 2014 to 2019, he was a Principal Investigator of the ERC Advanced
Grant ALEXANDRIA, where he worked on foundations for temporal search,
exploration, and analytics in web archives. He has published more than
440 scientific articles listed at DBLP, with an H-index (based on Google
Scholar) of 78. His research interests include information retrieval/search,
web science, artificial intelligence, social and semantic web, digital libraries,
and technology enhanced learning.

\end{IEEEbiography}

\begin{IEEEbiography}[{\includegraphics[width=1in,height=1.25in,clip,keepaspectratio]{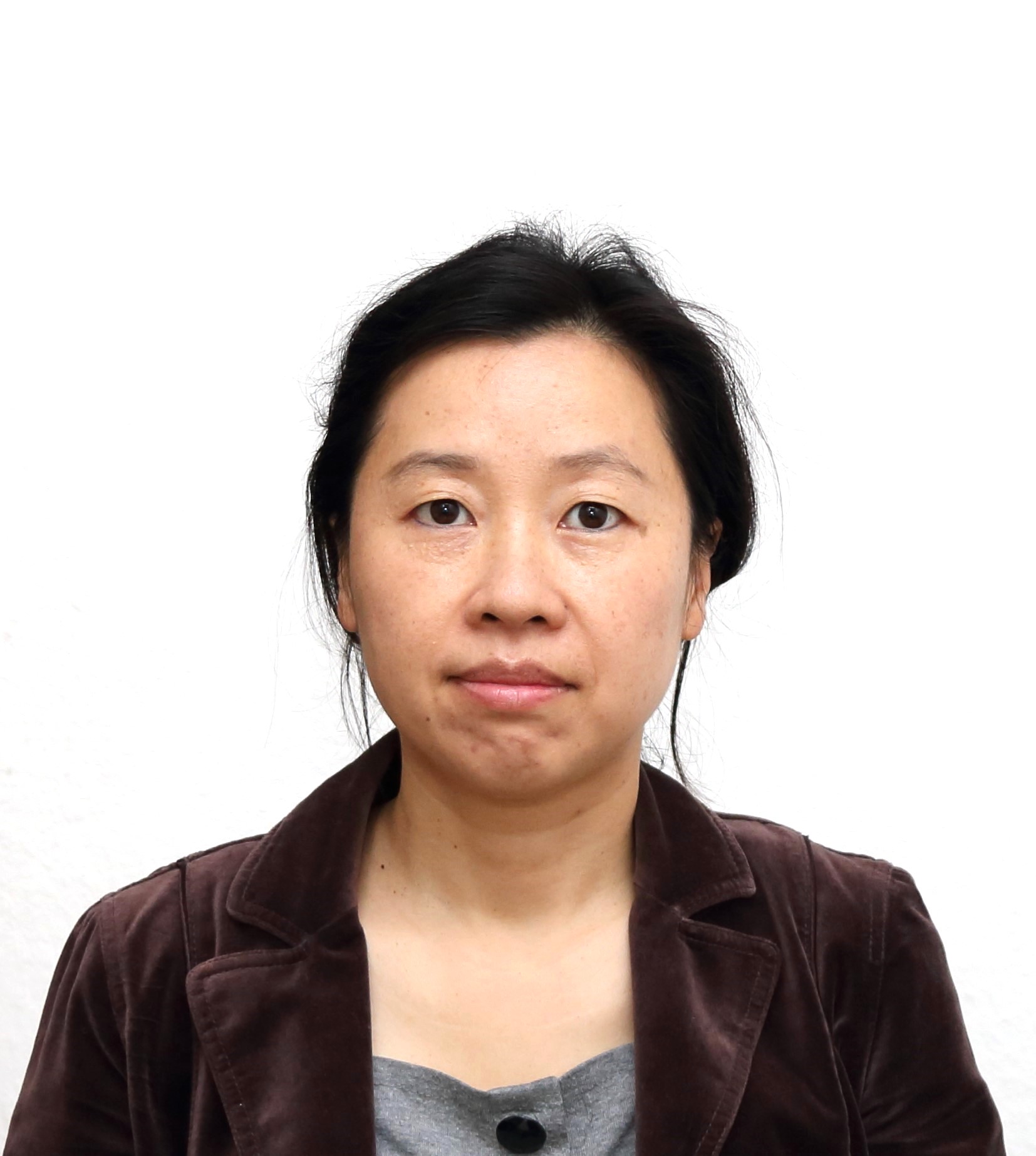}}]{Ming Tang} received the M.Sc. degree in informatics from the Free University of Berlin, Germany, and the Ph.D. degree in systems biology from the Johns Hopkins University School of Medicine, USA. She is currently a Group Leader with the L3S Research Center, Faculty of Electrical Engineering and Computer Science, Leibniz University Hannover. Her research interests include biomedical foundation models and agentic AI for biomedicine. She is involved in the CAIMed project (AI and Causal Methods in Medicine) and was previously involved in the International Future Lab for Artificial Intelligence, funded by the German Federal Ministry of Education and Research (BMBF).
\end{IEEEbiography}

\newpage

\EOD


\begin{thebibliography}{00}

\bibitem{ref_khawaji2025}
Khawaji, A., Martin, R.J. and Daspin, D., 2025. High-Resolution Medical Image Generation With Leak-Prevention Mechanism Using Quantum Transformer Learning. IEEE Access, 13, pp.206381-206401.


\bibitem{ref_vit}
Dosovitskiy, A., 2020. An image is worth 16x16 words: Transformers for image recognition at scale. arXiv preprint arXiv:2010.11929.

\bibitem{ref_transformer}
Vaswani, A., Shazeer, N., Parmar, N., Uszkoreit, J., Jones, L., Gomez, A.N., Kaiser, Ł. and Polosukhin, I., 2017. Attention is all you need. Advances in neural information processing systems, 30.

\bibitem{ref_saliency}
Simonyan, K., Vedaldi, A. and Zisserman, A., 2013. Deep inside convolutional networks: Visualising image classification models and saliency maps. arXiv preprint arXiv:1312.6034.

\bibitem{ref_ig}
Sundararajan, M., Taly, A. and Yan, Q., 2017, July. Axiomatic attribution for deep networks. In International conference on machine learning (pp. 3319-3328). PMLR.

\bibitem{ref_smoothgrad}
Smilkov, D., Thorat, N., Kim, B., Viégas, F. and Wattenberg, M., 2017. Smoothgrad: removing noise by adding noise. arXiv preprint arXiv:1706.03825.

\bibitem{ref_grad_cam}
Selvaraju, R.R., Cogswell, M., Das, A., Vedantam, R., Parikh, D. and Batra, D., 2017. Grad-cam: Visual explanations from deep networks via gradient-based localization. In Proceedings of the IEEE international conference on computer vision (pp. 618-626).

\bibitem{ref_legrad}
Bousselham, W., Boggust, A., Chaybouti, S., Strobelt, H. and Kuehne, H., 2025. Legrad: An explainability method for vision transformers via feature formation sensitivity. In Proceedings of the IEEE/CVF International Conference on Computer Vision (pp. 20336-20345).

\bibitem{ref_occlusion}
Zeiler, M.D. and Fergus, R., 2014, September. Visualizing and understanding convolutional networks. In European conference on computer vision (pp. 818-833). Cham: Springer International Publishing.

\bibitem{ref_rise}
Petsiuk, V., Das, A. and Saenko, K., 2018. Rise: Randomized input sampling for explanation of black-box models. arXiv preprint arXiv:1806.07421.
\bibitem{ref_pert_1}
Fong, R.C. and Vedaldi, A., 2017. Interpretable explanations of black boxes by meaningful perturbation. In Proceedings of the IEEE international conference on computer vision (pp. 3429-3437).

\bibitem{ref_pert_2}
Nalmpantis, A., Panagiotopoulos, A., Gkountouras, J., Papakostas, K. and Aziz, W., 2023. Vision diffmask: Faithful interpretation of vision transformers with differentiable patch masking. In Proceedings of the IEEE/CVF Conference on Computer Vision and Pattern Recognition (pp. 3756-3763).


\bibitem{ref_attn_not_expl_1}
Jain, S. and Wallace, B.C., 2019, June. Attention is not explanation. In Proceedings of the 2019 Conference of the North American Chapter of the Association for Computational Linguistics: Human Language Technologies, Volume 1 (Long and Short Papers) (pp. 3543-3556).

\bibitem{ref_attn_not_expl_2}
Serrano, S. and Smith, N.A., 2019, July. Is attention interpretable?. In Proceedings of the 57th annual meeting of the association for computational linguistics (pp. 2931-2951).

\bibitem{ref_att_rollout}
Abnar, S. and Zuidema, W., 2020. Quantifying attention flow in transformers. arXiv preprint arXiv:2005.00928.


\bibitem{ref_re_0}
Bach, S., Binder, A., Montavon, G., Klauschen, F., Müller, K.R. and Samek, W., 2015. On pixel-wise explanations for non-linear classifier decisions by layer-wise relevance propagation. PloS one, 10(7), p.e0130140.

\bibitem{ref_re_1}
Chefer, H., Gur, S. and Wolf, L., 2021. Transformer interpretability beyond attention visualization. In Proceedings of the IEEE/CVF conference on computer vision and pattern recognition (pp. 782-791).

\bibitem{ref_re_2}
Achtibat, R., Hatefi, S.M.V., Dreyer, M., Jain, A., Wiegand, T., Lapuschkin, S. and Samek, W., 2024. Attnlrp: attention-aware layer-wise relevance propagation for transformers. arXiv preprint arXiv:2402.05602.

\bibitem{ref_ia_1}
Qiang, Y., Li, C., Khanduri, P. and Zhu, D., 2023. Interpretability-aware vision transformer. arXiv preprint arXiv:2309.08035.

\bibitem{ref_ia_2}
Jeanneret, G., Simon, L. and Jurie, F., 2025. Disentangling Visual Transformers: Patch-level Interpretability for Image Classification. In Proceedings of the Computer Vision and Pattern Recognition Conference (pp. 2679-2689).

\bibitem{ref_ia_3}
Chowdhury, A., Paul, D., Mai, Z., Gu, J., Zhang, Z., Mehrab, K.S., Campolongo, E.G., Rubenstein, D., Stewart, C.V., Karpatne, A. and Berger-Wolf, T., 2025. Prompt-CAM: Making Vision Transformers Interpretable for Fine-Grained Analysis. In Proceedings of the Computer Vision and Pattern Recognition Conference (pp. 4375-4385).

\bibitem{vae}
Diederik, P.K. and Max, W., 2019. An introduction to variational autoencoders. Foundations and Trends® in Machine Learning, 12(4), pp.307-392.

\bibitem{c_vae}
Sohn, K., Lee, H. and Yan, X., 2015. Learning structured output representation using deep conditional generative models. Advances in neural information processing systems, 28.

\bibitem{dataset_brain}
Msoud Nickparvar. (2021). Brain Tumor MRI Dataset [Data set]. Kaggle. https://doi.org/10.34740/KAGGLE/DSV/2645886


\bibitem{dataset_COVID_1}
M.E.H. Chowdhury, T. Rahman, A. Khandakar, R. Mazhar, M.A. Kadir, Z.B. Mahbub, K.R. Islam, M.S. Khan, A. Iqbal, N. Al-Emadi, M.B.I. Reaz, M. T. Islam, “Can AI help in screening Viral and COVID-19 pneumonia?” IEEE Access, Vol. 8, 2020, pp. 132665 - 132676.
\bibitem{dataset_COVID_2}
Rahman, T., Khandakar, A., Qiblawey, Y., Tahir, A., Kiranyaz, S., Kashem, S.B.A., Islam, M.T., Maadeed, S.A., Zughaier, S.M., Khan, M.S. and Chowdhury, M.E., 2020. Exploring the Effect of Image Enhancement Techniques on COVID-19 Detection using Chest X-ray Images. arXiv preprint arXiv:2012.02238.

\bibitem{dataset_chest}
Kermany, Daniel; Zhang, Kang; Goldbaum, Michael (2018), “Large Dataset of Labeled Optical Coherence Tomography (OCT) and Chest X-Ray Images”, Mendeley Data, V3, doi: 10.17632/rscbjbr9sj.3

\bibitem{dataset_breast}
Al-Dhabyani, W., Gomaa, M., Khaled, H. and Fahmy, A., 2020. Dataset of breast ultrasound images. Data in brief, 28, p.104863.



\bibitem{ref_lime}
Ribeiro, M.T., Singh, S. and Guestrin, C., 2016, August. " Why should i trust you?" Explaining the predictions of any classifier. In Proceedings of the 22nd ACM SIGKDD international conference on knowledge discovery and data mining (pp. 1135-1144).

\bibitem{ref_shap}
Lundberg, S.M. and Lee, S.I., 2017. A unified approach to interpreting model predictions. Advances in neural information processing systems, 30.


\bibitem{ref_bagnet}
Brendel, W. and Bethge, M., 2019. Approximating cnns with bag-of-local-features models works surprisingly well on imagenet. arXiv preprint arXiv:1904.00760.


\bibitem{ref_nam}
Agarwal, R., Melnick, L., Frosst, N., Zhang, X., Lengerich, B., Caruana, R. and Hinton, G.E., 2021. Neural additive models: Interpretable machine learning with neural nets. Advances in neural information processing systems, 34, pp.4699-4711.


\bibitem{ref_sop}
You, W., Qu, H., Gatti, M., Jain, B. and Wong, E., 2025, October. Sum-of-parts: Self-attributing neural networks with end-to-end learning of feature groups. In Forty-second International Conference on Machine Learning.


\bibitem{ref_cls_patch}
Marouani, A., Siméoni, O., Jégou, H., Bojanowski, P. and Vo, H.V., 2026. Revisiting [cls] and patch token interaction in vision transformers. arXiv preprint arXiv:2602.08626.

\bibitem{ref_register}
Darcet, T., Oquab, M., Mairal, J. and Bojanowski, P., 2024, May. Vision transformers need registers. In International conference on learning representations (Vol. 2024, pp. 2632-2652).

\bibitem{ref_medical_patch_selection}
Nilsson, A. and Azizpour, H., 2024, July. Regularizing and Interpreting Vision Transformer by Patch Selection on Echocardiography Data. In CHIL (pp. 155-168).




\bibitem{ref_shahin}
Shahin, A., 2025. Fine-tuned ResNet34 for efficient brain tumor classification. Scientific Reports, 15(1), p.36910.
\bibitem{ref_neuronet}
Haque, R., Hassan, M.M., Bairagi, A.K. and Shariful Islam, S.M., 2024. NeuroNet19: an explainable deep neural network model for the classification of brain tumors using magnetic resonance imaging data. Scientific reports, 14(1), p.1524.
\bibitem{ref_disci}
Disci, R., Gurcan, F. and Soylu, A., 2025. Advanced brain tumor classification in MR images using transfer learning and pre-trained deep CNN models. Cancers, 17(1), p.121.



\end{thebibliography}
\end{document}